# The Use of Synthetic Data to Train AI Models: Opportunities and Risks for Sustainable Development

*Tshilidzi Marwala\*, Eleonore Fournier-Tombs\*\*, Serge Stinckwich\*\*\**
*\* United Nations University, Tokyo, Japan*
*\*\* UNU Centre for Policy Research, New York, USA*
*\*\*\* UNU Macau, Macau SAR, China*

**Abstract:**
In the current data driven era, synthetic data, artificially generated data that resembles the characteristics of real world data without containing actual personal information, is gaining prominence. This is due to its potential to safeguard privacy, increase the availability of data for research, and reduce bias in machine learning models. This paper investigates the policies governing the creation, utilization, and dissemination of synthetic data. Synthetic data can be a powerful instrument for protecting the privacy of individuals, but it also presents challenges, such as ensuring its quality and authenticity. A well crafted synthetic data policy must strike a balance between privacy concerns and the utility of data, ensuring that it can be utilized effectively without compromising ethical or legal standards. Organizations and institutions must develop standardized guidelines and best practices in order to capitalize on the benefits of synthetic data while addressing its inherent challenges.

**Recommended technical actions:**
- Use diverse data sources when creating synthetic datasets
- Use different types of generative AI models to create synthetic datasets
- Disclose or watermark all synthetic data and its provenance
- Calculate and disclose quality metrics for synthetic data
- Develop cybersecurity protocols to protect synthetic data and its source
- Prioritise non-synthetic data if possible

**Recommended policy actions:**
- Link synthetic data to global AI governance efforts
- Recognise synthetic data as a critical and unique issue in global data governance
- Establish global quality standards and security measures
- Promote global research networks on the safe and ethical use of synthetic data
- Clarify ethical guidelines, including transparency

**Introduction**

Using synthetic or artificially generated data in training AI algorithms is a burgeoning practice with significant potential. It can address data scarcity, privacy, and bias issues and raise concerns about data quality, security, and ethical implications. This issue is heightened in the global South, where data scarcity is much more severe than in the global North. Synthetic data, therefore, addresses the problem of missing data, leading, in the best case, to better representation of populations in datasets and more equitable outcomes. However, we cannot consider synthetic data to be better or even equivalent to actual data from the physical world. In fact, there are many risks to using synthetic data, including cybersecurity risks, bias propagation, and simply an increase in model error. This policy brief proposes recommendations for the responsible use of synthetic data in AI training and the associated guidelines to regulate the use of synthetic data. The objective of this policy brief is to explore the potential of synthetic data to accelerate the attainment of the SDGs through AI in the Global South while mitigating its important risks.

**Policy context**

As the multilateral system begins to turn towards the global governance of artificial intelligence, an important challenge will be addressing the growing digital divide. Today, in an era of accelerating digitization, the digital divide manifests itself not only in terms of Internet connectivity, but also, increasingly, in terms of representation and heterogeneity of datasets[i].

Data is critical at all stages of artificial intelligence development, especially during the training and testing phases. AI models trained on datasets representing only a segment of the population risk much higher rates for those not represented. In fact, many of the risks of artificial intelligence articulated in recent years have been due to homogeneity of datasets within and between countries.

A well-documented and glaring example was unearthed by Buolamwini and Gebru, who found that facial recognition systems used in the United States were trained on a very narrow dataset and had overwhelmingly high error rates for women of colour[ii]. There have also been numerous examples of women being stereotyped or discriminated against in AI systems for similar reasons, notably in human resources and financial applications.

But the problem of homogeneity in AI data is even more glaring in the Global South. Local subtleties, from the use of local languages, to clothing, to many other social, cultural and economic dimensions are still poorly represented in datasets. This is increasingly leading to effects on sustainable development, from poorer performance of AI systems in health, education and government services; to longer term effects on peace and governance.

A powerful solution proposed has been the use of synthetic data. Synthetic data, which is artificially created, can fill the gaps in homogenous datasets to increase diversity in data, with the objective of reducing discrimination of AI systems against specific groups. This type of technique has been praised as a quick way to rebalance biased datasets to increase the appropriateness of AI systems in the Global South. In fact, Gartner has argued that 60% of the data used for AI systems will be synthetically generated, as soon as 2024[iii].

## What is synthetic data?

Synthetic Data (SD) is information created by computer simulations or algorithms that reproduce some structural and statistical properties of real-world data. Data produced by this "synthesis" process can be images, videos, text or tabular data. SD are generally produced by a generative model, based on ground truth (domain knowledge, scientific theories or collected data), that will produce new samples of synthetic data.

It is important to note here that large language models (LLMs) production are also synthetic data. Therefore, the LLMs' accuracy in low-resource languages, such as TshiVenda, is less than in high-resource languages, such as English.

Synthetic data is increasingly being used to train AI algorithms, especially when real data is sensitive, scarce, or biased[iv].

There are various ways of creating synthetic data. The earliest forms of synthetic data generators are missing data imputation techniques[v][vi][vii]. In 1993[viii], Rubin came up with the idea of producing synthetic data based on statistical analysis that simultaneously allows the preservation of confidentiality of microdata.

With artificial intelligence advancements, Deep Generative Models (DGMs), based on neural networks have become the most preferred technique for generating synthetic data. Unlike neural networks that classify inputs, they are specialised in generating new outputs. They learn to generate new examples that are "similar" to existing collected data. Such models can produce high-fidelity images, music generation, sensory data for autonomous cars or robots, patient electronic health records or human mobility patterns in cities. Various DGMs approaches can be used like Variational Autoencoder (VAE), Generative Adversarial Networks and more recently Large Language Models (LLMs). Choices between them depends on specific use cases and data characteristics.

## Why use synthetic data?

Synthetic data offer numerous opportunities, such as rebalancing biased datasets, protecting data privacy, and reducing the cost of data collection. These are summarised in Table 1.

**Data availability**: Synthetic data can overcome limitations associated with data scarcity, enabling more robust AI training and development. Synthetic data generated using large language models have been used to train AI models for tasks such as disease diagnosis and developing new treatments (SDG3). In the financial services industry (SDG8), synthetic data has been used to train AI models for economic forecasting, fraud detection and risk assessment, among other tasks[ix]. In the climate science industry (SDG13), synthetic data is used to train AI models for weather forecasting and climate modeling, among other uses. This is essential for developing new mitigation and adaptation strategies for climate change.

Table 1 Synthetic data characteristics.

| Use of synthetic data | Description |
| --- | --- |
| Data availability | Synthetic data can address data availability and representation concerns by "completing" training datasets for AI systems. |
| Privacy protection | Synthetic data does not represent actual people, and so does not contain any personally identifiable information that could harm them in the case of a breach. |
| Bias reduction | Synthetic data can address imbalanced training datasets leading to AI bias, as in the case of gender bias, or ethnic bias. |
| Compliance | Synthetic data can be ethnic to train AI models when real data is restricted, such as in the medical field. |
| Cost | There can be cost benefits of using synthetic data instead of real data collection, although computational and environmental costs can still be important. |

**Privacy protection**: Synthetic data does not contain personally identifiable information (PII), making it a valuable tool for complying with data protection regulations and protecting user privacy[x]. In the healthcare industry, PII is removed or de-identified before using real-world healthcare data to generate synthetic data, allowing AI models to be trained on realistic data while protecting patients' privacy (SDG3)[xi].

**Bias reduction**: Synthetic data can be designed to be balanced and representative, helping to reduce bias in AI models. For example, synthetic data has been used to ensure that gender discrimination is minimized in artificial intelligence models, thereby advancing SDG5[xii].

**Compliance**: Synthetic data can be used to comply with regulations that restrict the use of real-world data. For instance, synthetic data can be utilized to train machine learning models without accessing sensitive data. Synthetic data has been used to generate sensitive medical images that are used for training medical students thereby advancing SDG4[xiii].

**Cost**: The generation of synthetic data can be more cost-effective than the collection of real-world data. This can be significant for applications requiring costly data collection, such as clinical trials and market research. However, the computational and environmental costs of generating synthetic data are substantial in some cases. Therefore, it is essential to consider this when choosing the generative model to synthesize data to advance SDG13.

## Main risks of synthetic data

Nevertheless, there are many risks to using synthetic data, such as data quality, cybersecurity, misuse, bias propagation, IP infringement, data pollution and data contamination.

**Data quality:** The quality and realism of synthetic data are crucial for effective AI training. Poorly generated synthetic data can lead to inaccurate and unreliable AI models. Depending on the methodology employed, the integrity of synthetic data can vary. Frequently, the data generated by a generative adversarial network (GAN) are highly realistic, but it can be challenging to control their distribution. Statistical models can generate more evenly distributed data, but the resulting data may be less plausible. A number of variables can impact the integrity of synthetic data. Firstly, the method used to generate synthetic data can substantially affect its quality. GAN-generated data are often more realistic than data generated by statistical models, but their distribution can be more challenging to control. Second, when more data is used to train the model, the quality of the synthetic data will be enhanced. This is because the model will have more data from which to learn and generate more realistic synthetic data. Third, if the quality of the real-world data used to train the model is high, the quality of the synthetic data will also be high because the model can learn from real-world data and generate similar synthetic data.

**Security risks**: Synthetic data, if reverse-engineered, could potentially reveal information about the underlying real data or the process used to generate it, posing security risks. Re-identification is therefore a real risk for synthetic data, especially if the source data used is published with the synthetic data, or if the model used to create the synthetic data "overfits" the training data, meaning that it too closely resembles the original dataset.

**Misuse:** The use of synthetic data in AI training raises ethical questions, such as the potential for misuse in creating deepfakes, or other deceptive AI technologies. Synthetic data has also been increasingly found to have intellectual property risks, especially when generating images from artistic source materials, or from other sources where human beings would have intellectual property.

**Bias propagation, data pollution or data contamination**: If the synthetic data is not balanced, misrepresents a population group, or is otherwise biased, its biases could propagate throughout trained models and even to other synthetic datasets. Synthetic data is generated from a dataset, and if that dataset is narrow, then that narrowness is projected into the synthetic dataset as identified by Segal et al. in 2021 for the case of X-ray images.[xiv] As the use of synthetic data democratised and became cheaper, more and more data available on the Internet will been generated by Deep Generative Models, and these inputs wil used again to train AI systems in a kind of negative feedback, where DGMs might collapse[xv]. This will become more and more difficult to separate what is synthetic data from what is real data at the end.

## Technical Recommendations

Here are some technical suggestions for utilizing synthetic data:

1. **Use diverse data sources when creating synthetic datasets:** When generating synthetic data, it is crucial to utilize a variety of data sources to ensure that the data are as diverse and has many independent characteristics as feasible. This may involve the use of both collected, real-world data and data from other sources, such as simulations, expert knowledge or participatory data from citizens[xvi].
2. **Use different types of generative AI models to create synthetic datasets:** It is essential to select a model that is suited to the task at hand and will generate data that is realistic and representative of the actual world.
3. **Disclose or watermark all synthetic data and its provenance:** It is essential to disclose where all synthetic data comes from and how it was produced[xvii], and comply with any Intellectual Property protection provision.
4. **Calculate and disclose quality metrics for synthetic data:** Once the synthetic data has been generated, its quality must be evaluated to ensure that it is suitable for the intended application[xviii]. This may involve examining the information for accuracy, completeness, and diversity.
5. **Prioritise non-synthetic data if possible:** Synthetic data can be a powerful resource, but it must be used responsibly. This requires being aware of the limitations of synthetic data and using it without misleading or deceiving the users of the synthetic data.

## Policy Recommendations

1. **Link Synthetic Data to Global AI Governance Efforts:** In ongoing Global AI Governance efforts, including those recommended by the United Nation's Multistakeholder Advisory Body on Artificial Intelligence, it will be essential to establish a working group on synthetic data to ensure that the risks outlined are addressed globally and comprehensively.
2. **Recognise Synthetic Data as a critical and unique issue in Global Data Governance**: Include questions related to synthetic data in parallel policy tracks on global data governance, to better prepare the international community for an increase in use in the next few years.
3. **Establish Global Quality Standards and Security Measures**: Develop and implement standards for generating and using synthetic data in AI training to ensure its quality and reliability. Implement robust security measures to prevent reverse-engineering of synthetic data and protect the integrity of AI training processes.
4. **Promote global research networks on the safe and ethical use of synthetic data:** Ensure that academic and policy research on synthetic data is well funded, and that researchers from the Global South and the Global North are able to collaborate effectively and exchange solutions.
5. **Clarify Ethical Guidelines, including Transparency**: Provide clear ethical guidelines for using synthetic data in AI training, addressing issues such as consent, transparency, and potential misuse. Organizations should be transparent about using synthetic data in AI training. This can build trust, ensure accountability, and foster responsible innovation.

**Conclusion**

The last year has seen a significant increase in the use of generative AI across sectors and skill level globally. These technologies have made the creation of synthetic datasets even more accessible than before. However, as outlined in this brief, inappropriate creation and use of synthetic datasets in artificial intelligence systems can have significant adverse consequences on sustainable development, especially by propagating AI bias. On the other hand, there are many cases in which the sound use of this type of data can bring benefits, such as enhancing medical research and reducing discriminatory outputs of artificial intelligence models. This brief therefore proposes an early step in standardising the use of synthetic data, by outlining both technical standards to be adopted by software developers and recommendations for policymakers. These efforts particularly aim to feed into current conversations on the global governance of artificial intelligence taking place at the United Nations and in other multilateral venues.

**Author bios:**
*Dr Tshilidzi Marwala* is the Rector of United Nations University, Headquartered in Tokyo, and Under-Secretary-General of the United Nations. He was previously the Vice-Chancellor and Principal of the University of Johannesburg. Marwala has published over 300 papers in peer-reviewed journals and conferences, 27 books on AI and related topics and holds five patents. He is a member of the American Academy of Arts and Sciences, the World Academy of Sciences (TWAS) and the African Academy of Science.

*Dr Eleonore Fournier-Tombs* is the Head of Anticipatory Action and Innovation at UNU Centre for Policy Research, focusing on developing methodological tools and policy recommendations related to AI and data at the United Nations. She is also an Adjunct Professor at the University of Ottawa Faculty of Law in Accountable AI and a Global Context, and a recurring lecturer on new technologies and cybersecurity for McGill University and Université de Montréal.

*Dr Serge Stinckwich* is a computer scientist and the Head of Research at United Nations University Institute in Macau, a UN think tank taking a human-centred lens to look at how we can amplify the positive contributions of digital technologies for sustainable development and mitigate their risks. His main research interests are in Modelling of Complex Systems, Social Simulation and the impact of Artificial Intelligence on the Sustainable Developement Goals (SDGs).

**Disclaimer:** The views and opinions expressed in this paper do not necessarily reflect the official policy or position of the United Nations University.

**REFERENCE**

[i] Fournier-Tombs, E. Local transplantation, adaptation and creation of AI models for public health policy. *Frontiers in Artificial Intelligence*, *6*, 1085671.

[ii] Buolamwini, J., & Gebru, T. (2018, January). Gender shades: Intersectional accuracy disparities in commercial gender classification. In *Conference on fairness, accountability and transparency* (pp. 77-91). PMLR.

[iii] White, A. (2021). By 2024, 60% of the data used for the development of AI and analytics projects will be synthetically generated. Gartner Blog. Accessed at: https://blogs.gartner.com/andrew_white/2021/07/24/by-2024-60-of-the-data-used-for-the-development-of-ai-and-analytics-projects-will-be-synthetically-generated/

[iv] Marwala, T. (2023). *Artificial Intelligence, Game Theory and Mechanism Design in Politics*. Springer Nature.

[v] Marwala, T. (Ed.). (2009). *Computational Intelligence for Missing Data Imputation, Estimation, and Management: Knowledge Optimization Techniques: Knowledge Optimization Techniques*. IGI Global.

[vi] Leke, C. A., & Marwala, T. (2019). *Deep learning and missing data in engineering systems* (p. 179). Berlin, Germany: Springer International Publishing.

[vii] Mbuvha, R., Adounkpe, J. Y., Houngnibo, M. C., Mongwe, W. T., Nikraftar, Z., Marwala, T., & Newlands, N. K. (2023). A novel workflow for streamflow prediction in the presence of missing gauge observations. *Environmental Data Science*, *2*, e23.

[viii] Rubin, D. B. (1993). Statistical disclosure limitation. *Journal of official Statistics*, *9*(2), 461-468.

[ix] Sidogi, T., Mongwe, W. T., Mbuvha, R., & Marwala, T. (2022, December). Creating Synthetic Volatility Surfaces using Generative Adversarial Networks with Static Arbitrage Loss Conditions. In *2022 IEEE Symposium Series on Computational Intelligence (SSCI)* (pp. 1423-1429). IEEE.

[x] Savage, N. (2023). Synthetic data could be better than real data. *Nature*. https://www.nature.com/articles/d41586-023-01445-8

[xi] Hernandez, M., Epelde, G., Alberdi, A., Cilla, R., & Rankin, D. (2022). Synthetic data generation for tabular health records: A systematic review. *Neurocomputing*, *493*, 28-45.

[xii] Sharma, S., Zhang, Y., Ríos Aliaga, J. M., Bouneffouf, D., Muthusamy, V., & Varshney, K. R. (2020, February). Data augmentation for discrimination prevention and bias disambiguation. In *Proceedings of the AAAI/ACM Conference on AI, Ethics, and Society* (pp. 358-364).

[xiii] Costello, J. P., Olivieri, L. J., Krieger, A., Thabit, O., Marshall, M. B., Yoo, S. J., Kim P.C., Jonas R. A., & Nath, D. S. (2014). Utilizing three-dimensional printing technology to assess the feasibility of high-fidelity synthetic ventricular septal defect models for simulation in medical education. *World Journal for Pediatric and Congenital Heart Surgery*, *5*(3), 421-426.

[xiv] Segal, B., Rubin, D. M., Rubin, G., & Pantanowitz, A. (2021). Evaluating the clinical realism of synthetic chest x-rays generated using progressively growing gans. *SN Computer Science*, *2*(4), 321.

[xv] Shumailov, I., Shumaylov, Z., Zhao, Y., Gal, Y., Papernot, N., & Anderson, R. (2023). The Curse of Recursion: Training on Generated Data Makes Models Forget. *arXiv preprint arxiv:2305.17493*.

[xvi] Tan, Y. R., Agrawal, A., Matsoso, M. P., Katz, R., Davis, S. L., Winkler, A. S., Huber A., Joshi A., El-Mohandes A., Mellado B., Mubaira C. A., Canlas F. C., Asiki G., Khosa H., Lazarus J. V., Choisy M., Recamonde-Mendoza M., Keiser O., Okwen P., English R., Stinckwich S., Kiwuwa-Muyingo S., Kutadza T., Sethi T., Mathaha T., Nguyen V.K, Gill A. & Yap, P. (2022). A call for citizen science in pandemic preparedness and response: beyond data collection. *BMJ Global Health*, *7*(6), e009389.

[xvii] Gebru, T., Morgenstern, J., Vecchione, B., Vaughan, J. W., Wallach, H., Iii, H. D., & Crawford, K. (2021). Datasheets for datasets. *Communications of the ACM*, *64*(12), 86-92.

[xviii] https://www.vanderschaar-lab.com/generating-and-evaluating-synthetic-data-a-two-sided-research-agenda/